\documentclass{article}

\usepackage{arxiv}

\usepackage{svg}
\usepackage{amssymb}
\usepackage{url}
\usepackage{microtype}
\usepackage{subcaption}

\usepackage{amsmath,amsfonts}
\usepackage{algorithmic}
\usepackage{array}
\usepackage{textcomp}
\usepackage{stfloats}
\usepackage{multirow}
\usepackage{verbatim}
\usepackage{graphicx}
\usepackage{chngcntr}
\newcommand\Tstrut{\rule{0pt}{2.0ex}}       

\def\BibTeX{{\rm B\kern-.05em{\sc i\kern-.025em b}\kern-.08em
    T\kern-.1667em\lower.7ex\hbox{E}\kern-.125emX}}
\usepackage{balance}

\usepackage[figuresright]{rotating}

\begin{document}


\title{Event-based Vision for Early Prediction of Manipulation Actions}


\author{
    Daniel~Deniz \\
    Computer Architecture and Technology \\
    CITIC, University of Granada\\
    Granada, Spain \\
    \texttt{danideniz@ugr.es} \\
\And
    Cornelia~Ferm\"uller \\
    Perception and Robotics Group \\
   University of Maryland \\ 
   Maryland, USA\\
    \texttt{fermulcm@umd.edu} \\
\And
    Eduardo~Ros \\
    Computer Architecture and Technology \\
    CITIC, University of Granada\\
    Granada, Spain \\
    \texttt{eros@ugr.es} \\
\And
    Manuel~Rodriguez-Alvarez \\
    Computer Architecture and Technology \\
    CITIC, University of Granada\\
    Granada, Spain \\
    \texttt{manolo@ugr.es} \\   
\And
    Francisco~Barranco \\
    Computer Architecture and Technology \\
    CITIC, University of Granada\\
    Granada, Spain \\
    \texttt{fbarranco@ugr.es} \\
}

\maketitle

\begin{abstract}
Neuromorphic visual sensors are artificial retinas that output sequences of asynchronous events when brightness changes occur in the scene. These sensors offer 
many advantages including very high temporal resolution, no motion blur and smart data compression ideal for real-time processing.
In this study, we introduce an event-based dataset on fine-grained manipulation actions and perform an experimental study on the use of transformers for action prediction with events. There is enormous interest in the fields of cognitive robotics and human-robot interaction on understanding and predicting human actions as early as possible. Early prediction allows anticipating complex stages for planning, enabling effective and real-time interaction. Our Transformer network uses events to predict manipulation actions as they occur, using online inference. The model succeeds at predicting actions early on, building up confidence over time and achieving state-of-the-art classification. Moreover, the attention-based transformer architecture allows us to study the role of the spatio-temporal patterns selected by the model. Our experiments show that the Transformer network captures action dynamic features outperforming video-based approaches and succeeding with scenarios where the differences between actions lie in very subtle cues. Finally, we release the new event dataset, which is the first in the literature for manipulation action recognition. Code will be available at \url{https://github.com/DaniDeniz/EventVisionTransformer}.
\end{abstract}

\keywords{event-based vision, online prediction, manipulation action prediction}



\section{Introduction}
\label{sec:introduction}

\begin{figure}[t]
  \centering
  \includegraphics[width=0.65\linewidth]{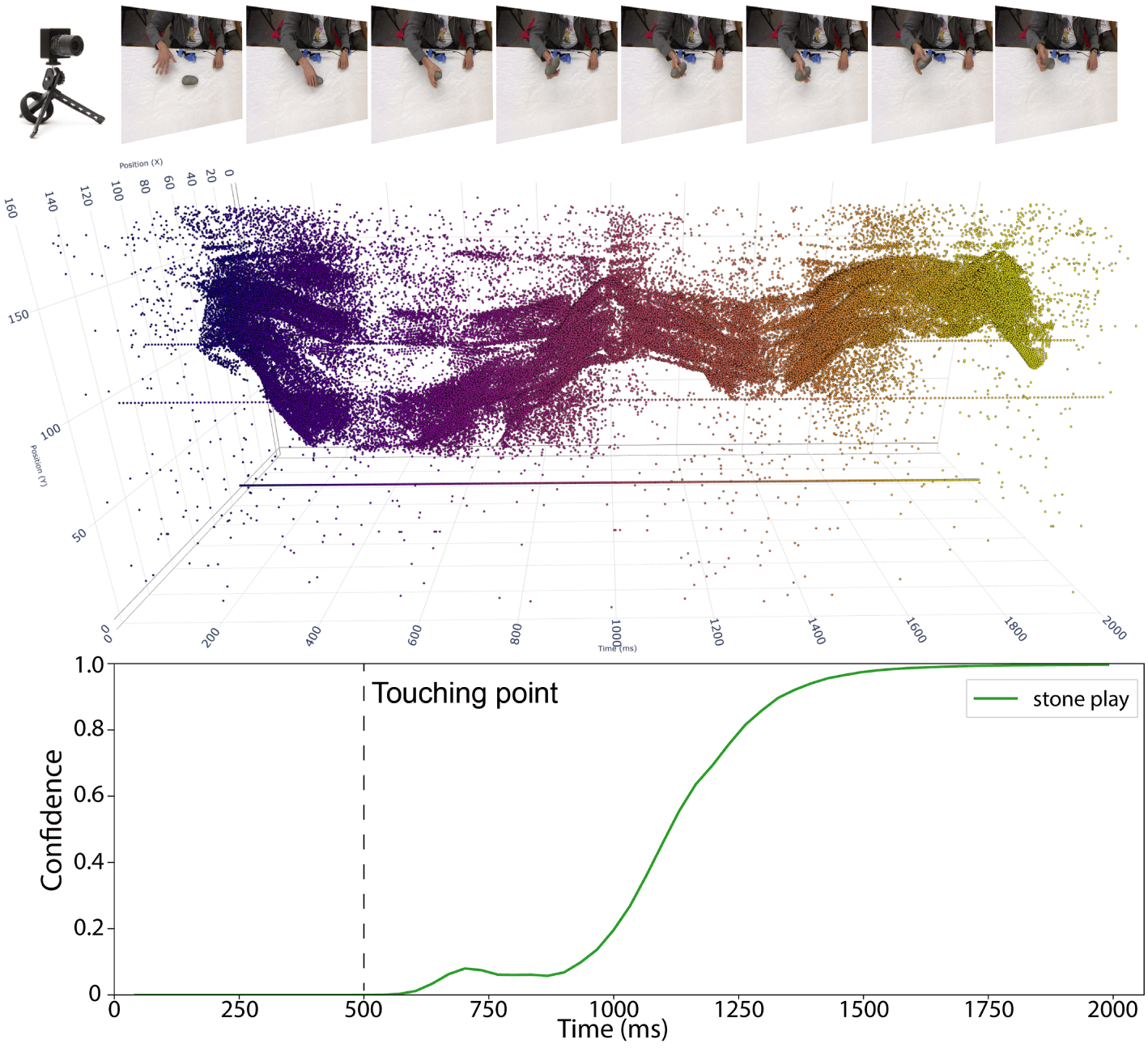}
  \caption{Example of online prediction of a manipulation action (``play with a stone'') using a Transformer architecture from a stream of asynchronous events. Event-wise processing provides a prediction with every new event. Top: action frames and continuous stream of events from the neuromorphic sensors. Bottom: online confidence prediction over time to classify the ongoing action early. In this case, about 0.7~s after the touching point (hand-object) the confidence for the (correct) action label already reaches approximately 0.8. Anticipation is crucial for human-robot interaction.}
  \label{figure:system_sample}
\end{figure}

Human Activity Recognition (HAR) aims at interpreting human activities by analyzing data from videos or wearables that embed inertial sensors. Advances in HAR have greatly contributed to the development of solutions for lifestyle monitoring \cite{deniz2020reconfigurable} and surveillance security \cite{roy2018suspicious, isern2020reconfigurable}. HAR is even more relevant for applications that involve robot-human interaction \cite{anagnostis2021human,yang2014cognitive,yang2015robot} -- a real-time task that requires anticipating 
the actions that humans are performing. 


Nowadays, state-of-the-art solutions for HAR use Deep Learning techniques. Most approaches use complex 2D and 3D Convolutional Neural Networks (CNNs) \cite{carreira2017quo, ddeniz2022efficient}, or combine 2D CNNs with recurrent neural networks (RNNs)~\cite{donahue2015long}. 
More recently, \textit{Transformers} have become very popular due to their great performance
\cite{vaswani2017attention}
in many fields 
including computer vision. For example, \cite{eusanio2020trans} proposes a combination of a ResNet network with a Transformer encoder module with multiple self-attention heads for hand gesture recognition, or \cite{mazzia2022action} proposes a Transformer that exploits human pose estimates for accurate and real-time activity recognition. Our work focuses on predicting manipulation actions, considering humans interacting with their hands with tools or objects to perform specific tasks.

For humans, action prediction is a continuous process that plays a crucial role, especially in collaborative tasks. In fact, humans constantly update their predictions based on new information, which modifies their belief of the ongoing action. 
Consequently, we consider manipulation action prediction as more than only assigning a label to a video. We consider that action prediction is not limited to the action itself but also to the prediction of its dynamics and effects while interacting with the objects or the environment. Moreover, since vision processing is computationally intensive, the delay to understand the action is unacceptable if classification takes place after completing the action. Reducing the latency means an advantage for planning ahead and acting, 
enabling closed perception-action loops in real time \cite{waltemate2016impact}, one of the main issues robotics has always struggled with \cite{corke1996dynamic}. Some works 
refer to ongoing prediction as \textit{online inference} that  provides predictions using only the historic information  \cite{chen2022gatehub}, different from conventional action classification, which works on the complete action segments. 

Neuromorphic visual sensors encode temporal luminance contrast, triggering asynchronous events at each pixel independently and with very-high temporal resolution. As a result, neuromorphic sensors smartly compress and reduce the  data to be processed only to changes on moving object contours and textures, and provide outputs with very low latency. 
These properties make event sensors the best fit for applications of visual motion, such as optical flow \cite{barranco2014contour}, tracking \cite{barranco2018real, gallego2017event}, motion segmentation \cite{mitrokhin2018event,mitrokhin2020learning, parameshwara20210}, or gait recognition \cite{wang2019ev}. In our case, online prediction is crucial for the development of low-latency predictive solutions that allow for updating the belief of the ongoing action as new asynchronous events come in. 

Using visual cues from the early stages of videos suffices for making a reasonably accurate prediction of the ongoing action \cite{ansuini2015intentions, fermuller2018prediction}. Studies suggest that observers catch subtle visual cues for recognition, such as the positioning of the fingers or the angles of adjacent fingers during the reaching phase \cite{ansuini2015predicting}. In this work, we argue that events or changes in the scene are sufficient to detect these cues.
There are notable differences between frame-based and event-based recognition. 
For example, since only motion is ``seen'', 
until the hand touches the object, only hand dynamics can be observed. 
Furthermore, while video-based approaches need to reconstruct action dynamics from the sequence of frames, neuromorphic solutions naturally focus on dynamic features. These are crucial for recognition when the differences between actions lie in very subtle cues, as in manipulation actions where subjects perform different actions with the same objects or tools.  

Deep learning methods are data hungry; training requires a large amount of diverse data which is costly due to the time required for collection and curation. Event cameras are still novel sensors that became popular around 2008, and thus there is  a lack of event datasets for specialized tasks. 
In this work, we release, to the best of our knowledge, the first event dataset 
for manipulation action classification, called \textit{Event-based Manipulation Action Dataset}\footnote{https://github.com/DaniDeniz/DavisHandDataset-Events}. \textit{E-MAD} was recorded during the \textit{Telluride Neuromorphic Cognition Engineering Workshop}\footnote{https://sites.google.com/view/telluride-2023/} and contains 750 samples of event sequences performing 30 manipulation actions using 6 different objects. It has recordings of actions with very subtle differences in the motion, thus, analyzing the dynamics of the movement of the hand and objects is crucial to distinguish between activities. 

We propose a Transformer encoder architecture that captures temporal dependencies by processing embedding tokens extracted by a 2D CNN network relying on self-attention (inspired by \cite{vaswani2017attention}). This end-to-end causal Transformer is adapted for online prediction and event-wise computation. One of the most important results is that the Transformer is able to successfully capture motion dynamics from actions. 



The main contributions of our paper are summarized next:
\begin{itemize}
    \item The novel Event-based Manipulation Action Dataset (\textit{E-MAD}) featuring subjects manipulating objects and performing different actions with them. The dataset is the event-based counterpart of the frame-based \textit{MAD} \cite{fermuller2018predictionurl}, and thus it can also be used in future comparisons between event- and frame-based approaches.
    
    \item An event-wise predictive \textit{Transformer} model for manipulation action recognition that updates its belief on the ongoing action with every new asynchronous event. The model is compared with the state of the art, including other predictive models and networks that require the full sequence of events  for classification.
    \begin{itemize}
        \item A study of the model generalization capacity. Also, a subject- and object-agnostic approach is introduced, which shows that cues only from the hand dynamics suffice for the recognition while considerably reducing the computations.
        \item A discussion on self-attention maps that illuminates which spatio-temporal patterns are key for the recognition. It is shown that there are significant differences between \textit{discrete} and \textit{cyclic} actions. 
    \end{itemize}
\end{itemize}

\section{Related Methods}
\label{sec:related}
This section briefly reviews the principles of event-based vision and then the literature on related tasks of action recognition using event sensors.
Finally, we describe event-based datasets for action classification. 

\textbf{Event generation}: Event cameras are bio-inspired sensors that trigger asynchronous events independently for each pixel when the brightness for that specific location changes beyond a given threshold. The output is a sequence of events where each event is represented as a tuple $e_i=\{u_i,t_i,p_i\}$ with $u_i=(x_i,y_i)$  the event coordinates, $t_i$ the time when the event occurred, and $p_i \in \{-1,1\}$ the polarity of the change (negative if the light decreases and positive if it increases). 
\textbf{Event-based Action Recognition}: Recently, several solutions have been presented for action recognition using event data. Many approaches bin events from small time intervals into images to apply conventional vision approaches, such as CNNs and Recurrent Neural Networks (RNNs) \cite{moreno2022visual}.
One of the main drawbacks of these approaches 
is the loss of temporal information, precisely, one of the main advantages of event data. Other works quantize events maintaining the temporal information. For example, in N-HAR \cite{pradhan2019n}, authors use memory surfaces built from sparse events as input to a deep CNN, specifically a variant of the Inception V3 architecture.
The study \cite{annamalai2022event} proposes an unsupervised task-agnostic LSTM architecture, called Event-LSTM, for converting events into time surfaces. The proposal ensures velocity invariance through an asynchronous 2D spatial grid sampling stage.


Another application is human gesture recognition, which is used in tasks such as the translation of sign language
and in human-computer interfaces.
These are very complex tasks where the motion dynamics over time are critical for recognizing the type of action. As pointed out in \cite{vasudevan2020introduction}, sign language and activity recognition usually involve visuospatial patterns at high speeds. Different approaches have been proposed in the literature. For example, \cite{vasudevan2020introduction} proposed two approaches based on biologically-inspired Spiking Neural Networks (SNNs) using backpropagation, namely SLAYER and Spatio Temporal Back Propagation (STBP). 
\cite{baldwin2021time} introduced Time-Ordered Recent Event (TORE) volumes to compactly store raw spike timing information, and evaluated them on various tasks, including gesture recognition. Finally, authors in \cite{sabater2022event} proposed an Event Transformer framework for gesture recognition that efficiently processes event-data using a patch-based event representation.
In their conclusions, the authors point out that approaches that avoid stacking events into frames and exploit the temporal information achieve the best results for gesture classification. 



\textbf{Manipulation action datasets}: There are several datasets on event-based action and gesture recognition, but not all are useful for evaluating manipulation actions. \textit{DVSACT16} \cite{hu2016dvs} is a dataset that contains event sequences of up to 50 actions (most of them sports), recorded with a \textit{DAVIS240C} event sensor viewing video sequences from the well-known \textit{UCF-50 Action Recognition} dataset \cite{reddy2013recognizing} displayed on a monitor. Although valuable for event-based action recognition, the monitor refresh rate adds unrealistic noise. Furthermore, most actions are full-body while in our task, we are focused on manipulations.

Closer related to our task is the \textit{DVS 128 Gesture Recognition} dataset of IBM \cite{amir2017low}. It contains 11 hand gestures from 29 different subjects and has more than 1300 recordings. However, it was designed for problems such as human-computer interaction and therefore does not include object manipulations. Interestingly, it includes different illumination conditions. Finally, the \textit{SL-Animals-DVS dataset} \cite{vasudevan2020introduction} has around 1100 samples from 58 subjects performing 19 gestures in Spanish sign language. Subjects perform different actions mainly with their hands. This makes recognition challenging because gestures are usually performed very fast, and the background scenes are diverse. 
For gesture recognition as well as for our problem, it is crucial to take into account the dynamics of the activities encoded by the events. However, different from gesture recognition, we are interested in manipulation of objects with the hands, for which there do not exist datasets yet.

\section{Our Approach}
\label{sec:our_approach}
Next we describe our algorithms for the prediction of manipulation actions using as input asynchronous events from a neuromorphic sensor. Our solution builds time surfaces from events, which are fed to a neural network architecture, each as a 2D grid.
To take advantage of the very accurate timing information of asynchronous events, the network architecture uses a transformer that captures long-range dependencies between events using self-attention. In the experiments, we compare the proposed event-based method with other \textit{classification} and \textit{prediction} architectures. 

\subsection{Feature extraction from event-based data}
\label{sec:event_processing}

To process asynchronous events $\textbf{e}_i$, we build \textit{Time Surfaces}, which allow us to use conventional algorithms while taking advantage of the benefits of event-based vision. The \textit{Time Surface} representation is a local descriptor common in event-based vision, which integrates the information of groups of events over time.

In our work, we use $\Gamma_{e}(\textbf{u},t)$ as described in Eq. \ref{equation:decaying_time_surface}, which is a \textit{Time Surface} built by summing events weighted with an exponentially decaying factor \cite{afshar2020event}. $\textbf{u} = (x_i,y_i)$ denotes the set of all event spatial coordinates and $t$ is the time when the surface is built. The function $\sum_{e}(u_i)$ maps event time to spatial coordinates, and $P_e(u_i)$ maps the polarity of each event. The decaying factor weighs the information provided by the events giving more importance to recent events and decaying earlier ones smoothly towards zero over time. 

\begin{equation}
\Gamma_{e}(\textbf{u},t)= \begin{cases}
P_{e}(u_i)\cdot e^{(\frac{\sum_{e}(u_i)-t}{\tau})}, & \sum_{e}(u_i)\leq t \\ 
0 & \sum_{e}(u_i) > t 
\end{cases}
\label{equation:decaying_time_surface}
\end{equation}

One of the challenges of working with event-vision sensors is dealing with noisy and spurious events \cite{gallego2020event}. Given the intrinsic sparsity of event-vision, it is crucial to reduce noise to extract meaningful information \cite{mitrokhin2018event}. The first mechanism for noise reduction consists of removing spatio-temporal isolated events,
using a salt-and-pepper-like filter.

\begin{figure*}[t]
  \centering
  \includegraphics[width=1\linewidth]{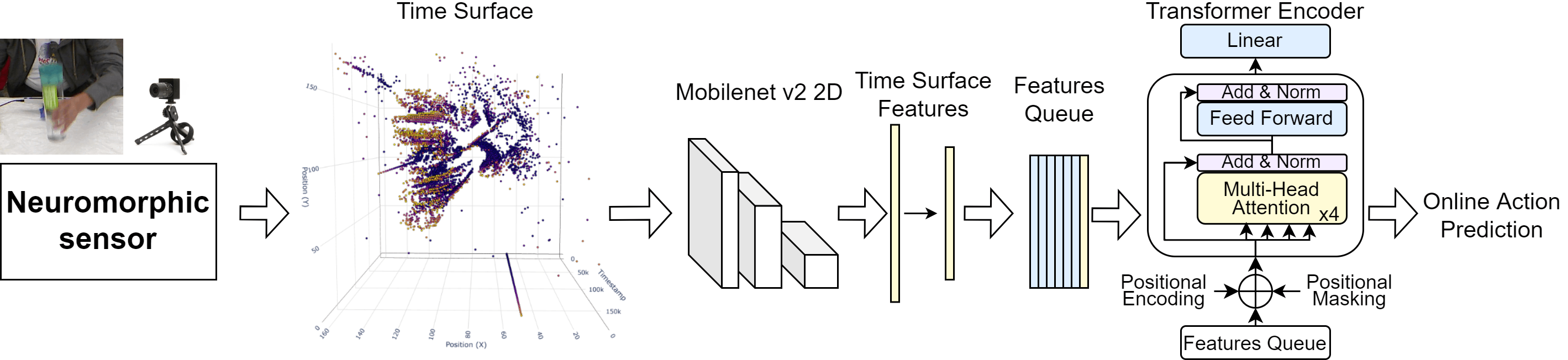}
  \caption{Pipeline of the proposed \textit{Mobilenet-based Transformer} architecture for online inference. The event camera triggers asynchronous events which are accumulated into a Time Surface with exponential decay \cite{afshar2020event}, in this example with a constant $\tau=33ms$. A \textit{Mobilenet} architecture extracts spatio-temporal features from the time surfaces.
 These features are projected into a 256-dimensional latent-space using dense layers. The features are queued and the network analyzes the temporal relationship between events in different time surfaces through self-attention to identify which action is being carried out. Note that feature extraction is run only once per time surface and the online prediction only needs to analyze new chunks of events integrating information in a causal manner i.e. using only previously processed time surfaces. This transformer with only one encoder layer and four causal self-attention heads captures long-range dependencies among events, which are embedded in the feature queue.}
  \label{figure:transformer_architecture}
\end{figure*}

\subsection{Action Recognition via Deep Learning}
\label{sec:action_recognition_dl}
We classify action recognition architectures into classification and predictive models. \textit{Classification models} require to be fed with the data
of the complete action. Since inference is performed with the whole data, either as frames or event temporal surfaces, this results in longer latency. On the other hand, \textit{predictive models} perform online prediction, integrating the information frame by frame or as in our case, event by event.
Online prediction systems offer benefits such as faster continuous prediction while reducing memory and computation costs \cite{kondratyuk2021movinets}. 


Our approach is based on a predictive continuous solution that uses a popular attention-based \textit{Transformer} model \cite{vaswani2017attention}. This sequence-modeling network is composed of a stack of identical transformer layers with attention heads. Each independent attention head produces a different attention distribution.
For our use case, only the encoder component of this highly paralellizable architecture
is used, which is sufficient  
to efficiently capture long-time range dependencies in sequences of events, and leads to interpretable online prediction. 

Our Transformer network analyzes every new time surface built from new incoming events. Specifically, a \textit{2D Mobilenet} model \cite{sandler2018mobilenetv2} extracts the spatio-temporal features from the temporal surfaces that are positionally encoded. Then, a single-layer Transformer encoder block analyzes the relationship between the new time surface and the previous ones. In this step, a look ahead mask is applied to force causal attention. 
In other words, the search is limited to previous temporal surfaces, never to future ones, ensuring causality. This way the model provides online prediction from continuous sequences of events. Finally, fully connected layers generate the manipulation action prediction.

To fully understand how the transformer model is converted into a predictive architecture, please refer to Figure \ref{figure:transformer_architecture}. The spatio-temporal features extracted by the \textit{Mobilenet 2D} architecture are queued, along with the previous spatio-temporal features from previous time surfaces. This stateful architecture preserves the features in a queue between inferences. Thus, every time a new time surface needs to be processed, its spatio-temporal features are queued and processed by the Transformer Block to learn the temporal relationship to the previous features.
By following this approach, spatio-temporal features are extracted from the time surfaces only once, which happens to be also the most computationally expensive part of the processing.

\subsection{Event training procedure}
\label{sec:event-training}
Firstly, event-data is processed following the approach for feature extraction described in Section \ref{sec:event_processing}. We project events into time surfaces.
A new time surface is built every 33 ms and thus, the system generates around 30 time surfaces per second.

For training, we feed our neural model with the time surfaces from a time span of two seconds (60 Time Surfaces). This is the average time required to capture the dynamics of the manipulation actions in the different datasets included in our evaluation. However, our predictive network architectures are designed for online prediction, 
to update the prediction on any new incoming time surface.

Our data augmentation
applies random rotations, cropping and horizontal flipping operations.
The spatial resolution of Time Surfaces is set to $144\times144$.

Training is done in two phases. First, we freeze the model backbone weights and train only the output layers for 60 epochs. Then the model is fine-tuned by retraining all layers for 60 more epochs. For our \textit{EMAD} dataset (see Section \ref{sec:manipulation_actions_recognition} for a detailed description), because of the limited number of samples per class (only 25), training is done in three stages. First, models are trained to distinguish between the 6 categories of object interactions for 60 epochs using a batch size of 8; second, weights are frozen and transfer learning is applied to identify the 30 manipulation actions by retraining only the final layers for 10 epochs. Finally, models are end-to-end fine-tuned for 100 additional epochs.

\section{Discussion and Results} 
In this section we describe the implementation details and evaluate our model against the state of the art on event-based datasets for action and gesture recognition. Next we describe our \textit{Event-based Manipulation Action Dataset}.
We analyze the model's self-attention mechanisms on event patterns. Finally, we compare our approach to similar frame-based approaches in accuracy and time performance.

\subsection{Event-based action recognition models}
\label{sec:models_discussion}
We have implemented and adapted four architectures, two classification and two predictive models. 

The two classification architectures are:  \textit{Mobilenet v2 3D} \cite{sandler2018mobilenetv2} and  \textit{Inception 3D} \cite{szegedy2015going}, both 3D inflated versions of their 2D convolutional variants based on 3D convolution operations capturing the spatio-temporal features of the input. The \textit{Mobilenet v2 3D} model has been designed to minimize the floating-point operations (FLOPS), different from high-end vision architectures such as \textit{Inception}. 
As mentioned, 
since these are not memory-based architectures, they require as input the complete sequence of frames or the whole stream of events at inference. 

The predictive models are: A \textit{Mobilenet LSTM} and our \textit{Mobilenet Transformer} \cite{vaswani2017attention}. The \textit{Mobilenet LSTM} is a Convolutional Recurrent Neural Network. A 2D Mobilenet backbone analyzes the time surfaces, then, a Long-short Term Memory (LSTM) \cite{hochreiter1997long} layer models the temporal features of the activity.  Our \textit{Mobilenet Transformer} has a 2D Mobilenet backbone and a Transformer encoder block (see Section \ref{sec:action_recognition_dl} for more details). As for the implementation details, we use a single Transformer encoder layer with 4 self-attention heads that process 256-dimensional features estimated from temporal surfaces constructed from packets of events. 
For the predictive versions, we introduce dense layers before the temporal analysis to project the features into a smaller latent dimension. This significantly reduces the computational complexity of the temporal analysis (with LSTM or Transformer blocks).

\subsection*{Comparison of event-based models}
Table \ref{table:eval_other_dataset} shows the accuracy performance for the \textit{DVS 128 Gesture Recognition} and the \textit{SL-Animals-DVS} datasets. For the \textit{DVS 128 Gesture Dataset} we used the train and test splits defined by the authors \cite{amir2017low} and for the \textit{SL-Animals-DVS dataset} we used a K-fold cross validation ($K=4$), following the same approach than in \cite{vasudevan2020introduction}. 

The first 4 rows summarize the results of state-of-the-art models using biologically-plausible solutions based on Spiking Neural Networks (SNNs). The performance values reported here are taken from their original papers. The next three rows show the results of the models described above, which are popular frame-based alternatives adapted to learn on temporal surfaces built from events. The last row shows the results from our Transformer architecture. 

For the DVS Gesture dataset all models have close to the state-of-the-art performance. For the SL-Animals-DVS dataset,
the models adapted from frame-based popular architectures reach much higher high accuracy performance than the SNN architectures. Our transformer-based approach reaches up to 36\% higher performance compared to the SLAYER approach \cite{vasudevan2020introduction}.

\begin{table}[t]
    \centering
    \caption{Accuracy of models for event-based action detection}
    \label{table:eval_other_dataset}
    \setlength{\tabcolsep}{1.8pt}
    \begin{tabular}{ l  r  r  r }
        \hline
        Architecture & DVS 128 Gesture & SL-Animals-DVS  \\
        \hline
        RG-CNN \cite{bi2020graph} & 97.20 & - \\
        SLAYER \cite{vasudevan2020introduction} & - & 60.09\\
        TORE \cite{baldwin2021time} & 96.20 & 85.10\\
        EvT \cite{sabater2022event} & 96.20 & 88.12\\
        
        \hline \Tstrut
        Inception 3D & \textbf{97.72} & \textbf{96.53}~$\pm$~0.85 \\ 
        Mobilenet v2 3D & 96.21 & $92.78\pm1.21$ \\
        Mobilenet LSTM & 96.21 & $94.47\pm0.61$ \\
        \hline \hline
        \textbf{Mobilenet Transf. (Ours)} & 96.21 & $94.11\pm0.59$ \\
        \hline
    \end{tabular}
\end{table}

\subsection{Dataset for manipulation action recognition}
\label{sec:manipulation_actions_recognition}
\begin{figure}[t]
    \centering
    \includegraphics[width=0.65\linewidth]{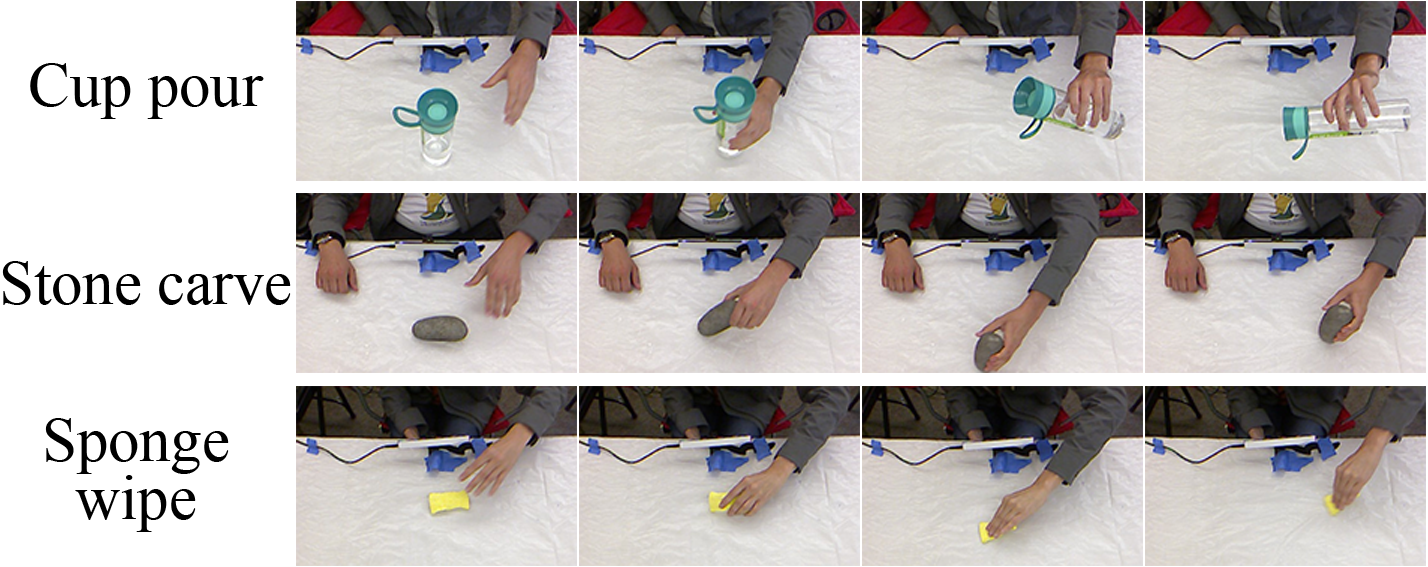}
    \caption{RGB frames from samples from the \textit{Manipulation Action Dataset} \cite{fermuller2018prediction}. It shows samples of \textit{pouring liquid into a cup}, \textit{carving with a stone}, and \textit{wiping with a sponge}.}
  \label{figure:samples_mad}
\end{figure}

Next we describe the \textit{Event-based Manipulation Action Dataset} (\textit{E-MAD}), the first dataset on manipulation actions using events. It is the event-based counterpart of the RGB video dataset (see Figure \ref{figure:samples_mad}) \textit{Manipulation Action Dataset} introduced in \cite{fermuller2018prediction, fermuller2018predictionurl}.

\textit{E-MAD} was recorded using a \textit{DAVIS240} neuromorphic sensor with 240x180 resolution. Subjects performed  manipulations on single objects while seated at a table.
All actions started with the subjects with their hands on the table (palms down). When the recording starts, the subject moves his/her hand to reach the object, performs the action, and then returns to the initial resting position. \textit{E-MAD} contains actions for  one more object than \textit{MAD}, namely the spatula.
We can roughly categorize the recorded actions into two kinds:
\textbf{cyclic} actions, which are repetitive and performed continuously, such as \textit{shaking a cup} or \textit{scraping with a spatula}, and \textbf{discrete} actions, done only once such as \textit{drinking from a cup} or \textit{scooping with a spoon} (see the discussion of results in section \ref{sec:self_attention}). 

The recordings are from five subjects (\textit{S1, S2, S3, S4, and S5}) with 750 samples of 30 manipulation actions using 6 different objects. The dataset is balanced: every manipulation action is represented with exactly 25 samples. The list of actions recorded with each object are as follows:
\begin{itemize}
    \item \textbf{Cup}: \textit{drink, pound, shake, move}, and \textit{pour}.
    \item \textbf{Stone}: \textit{pound, move, play, grind}, and \textit{carve}.
    \item \textbf{Sponge}: \textit{squeeze, flip, wash, wipe}, and \textit{scratch}.
    \item \textbf{Spoon}: \textit{scoop, stir, hit, eat}, and \textit{sprinkle}.
    \item \textbf{Knife}: \textit{cut, chop, poke a hole, peel}, and \textit{spread}.
    \item \textbf{Spatula}: \textit{flip, lift, cut, squeeze}, and \textit{scrape}.
\end{itemize}

\begin{table}[t]
    \centering
    \caption{Ablation study: Accuracy for actions on object classification ($N=6$)}
    \label{table:eval_six_objects}
    \begin{tabular}{ l r}
        \hline
        DL architecture & Accuracy  \\
        \hline \Tstrut
        Inception 3D &  91.98 $\pm$ 7.11\\ 
        Mobilenet v2 3D &  93.33 $\pm$ 3.59 \\
        Mobilenet LSTM & 83.60 $\pm$ 10.3 \\
        \textbf{Mob. Transf. (Ours)} & 91.46 $\pm$ 7.66 \\
        \hline
    \end{tabular}%
\end{table}

\begin{table}[t]
    \centering
    \caption{Ablation study: Accuracy for manipulation action classification ($N=30$)}
    \label{table:eval_manipulation_actions}
    \begin{tabular}{ l r }
        \hline
        DL architecture & Accuracy  \\
        \hline \Tstrut
        Inception 3D & 77.33 $\pm$ 7.14\\ 
        Mobilenet v2 3D &  80.53 $\pm$ 7.35\\
        Mobilenet LSTM & 81.19 $\pm$ 2.55\\
        \textbf{Mob. Transf. (Ours)} & 86.80 $\pm$ 4.04 \\
        \hline
    \end{tabular}%
\end{table}


\subsection{Prediction study}
In this section, we study the generalization of the classification over subjects. We first evaluate over the 
six action categories on the different objects (referred to as super-categories), then on the 30 different actions.
We trained 5 classifiers, leaving each time one subject out that was used only for testing purposes. For each classifier, the data was split into 85\% for training and 15\% for validation. The training procedure was explained in Section \ref{sec:event-training}.


Table \ref{table:eval_six_objects} shows the accuracy of the models for the classification of the 6 super-categories. The solutions reach a minimum average accuracy of about 83.60\% (\textit{Mobilenet LSTM}). In addition, most alternatives offer a recognition performance above 90\% with an average variation of 7-8\%. Hence, the architectures are capable of accurately identifying the objects that subjects manipulate. 

Table \ref{table:eval_manipulation_actions} shows the results for the classification on all 30 actions. We see that models based on 3D convolutions (\textit{Inception 3D, Mobilenet v2 3D}) in general have lower recognition performance with an average accuracy of about 80\%. For the worst case, the average accuracy of the most complex architecture, \textit{Inception 3D} is only $77.33\%\pm7.14$. 


Table \ref{table:eval_manipulation_actions} also shows the performance of the continuous predictive models.
The \textit{Mobilenet LSTM} model achieves an average accuracy of $81.19\%\pm2.55\%$ about a point higher than the best 3D-convolution-based model. But the highest average accuracy is obtained by the \textit{Mobilenet Transformer}, reaching up to $86.80\%\pm4.04\%$. Furthermore, note how this last model also tends to generalize better than the others, reaching the best recognition performance.  
This finding is relevant since online predictive models allow inference as soon as the first data arrives, and they improve over time, reducing the latency of prediction.

\subsection{Online prediction evaluation}
\begin{figure*}[t]
    \begin{center}
        \begin{minipage}[t]{0.5\textwidth}
	        \centering
 	        \includegraphics[width=\textwidth]{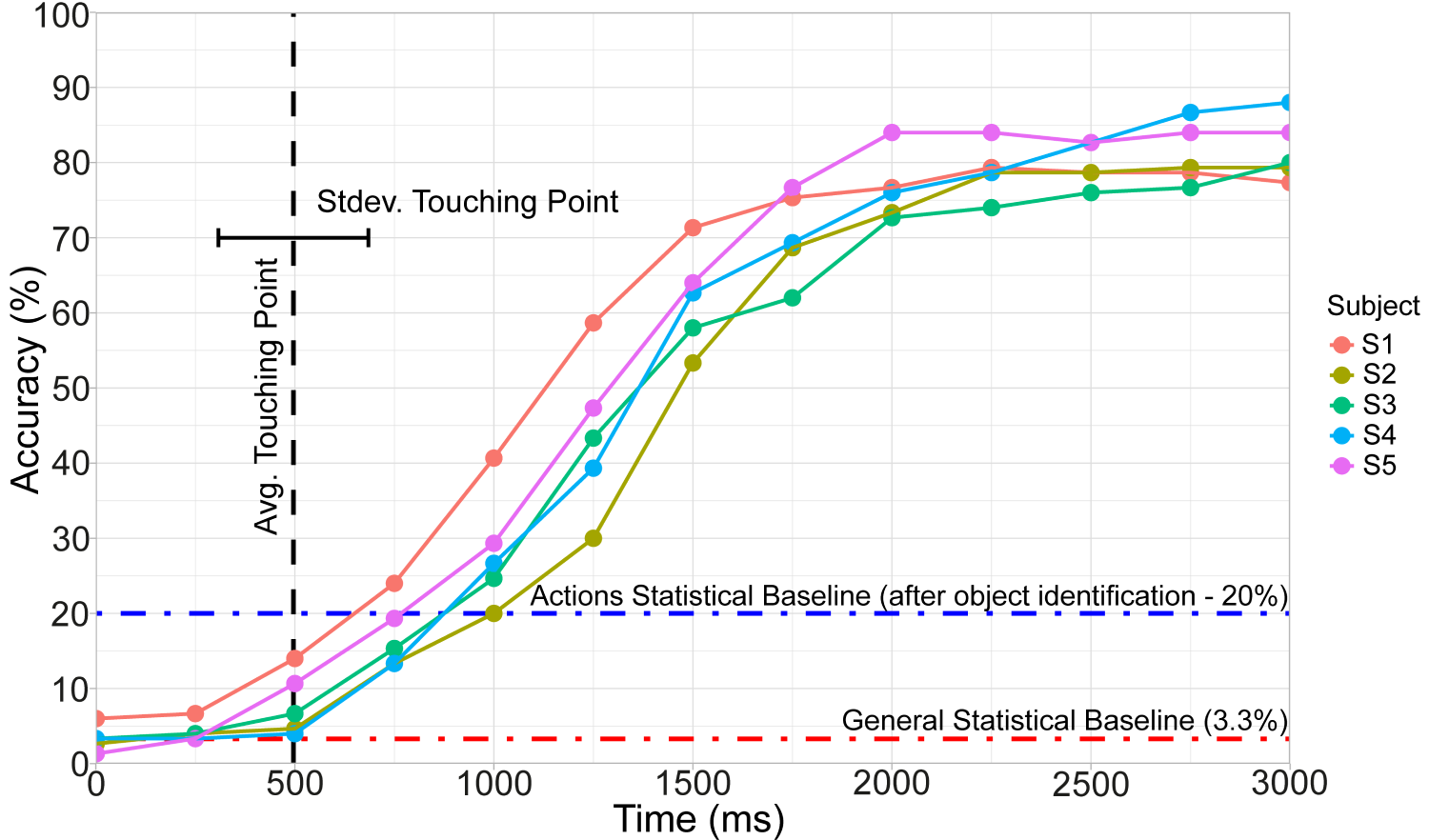}
        \end{minipage}%
        \begin{minipage}[t]{0.5\textwidth}
	        \centering
 	        \includegraphics[width=\textwidth]{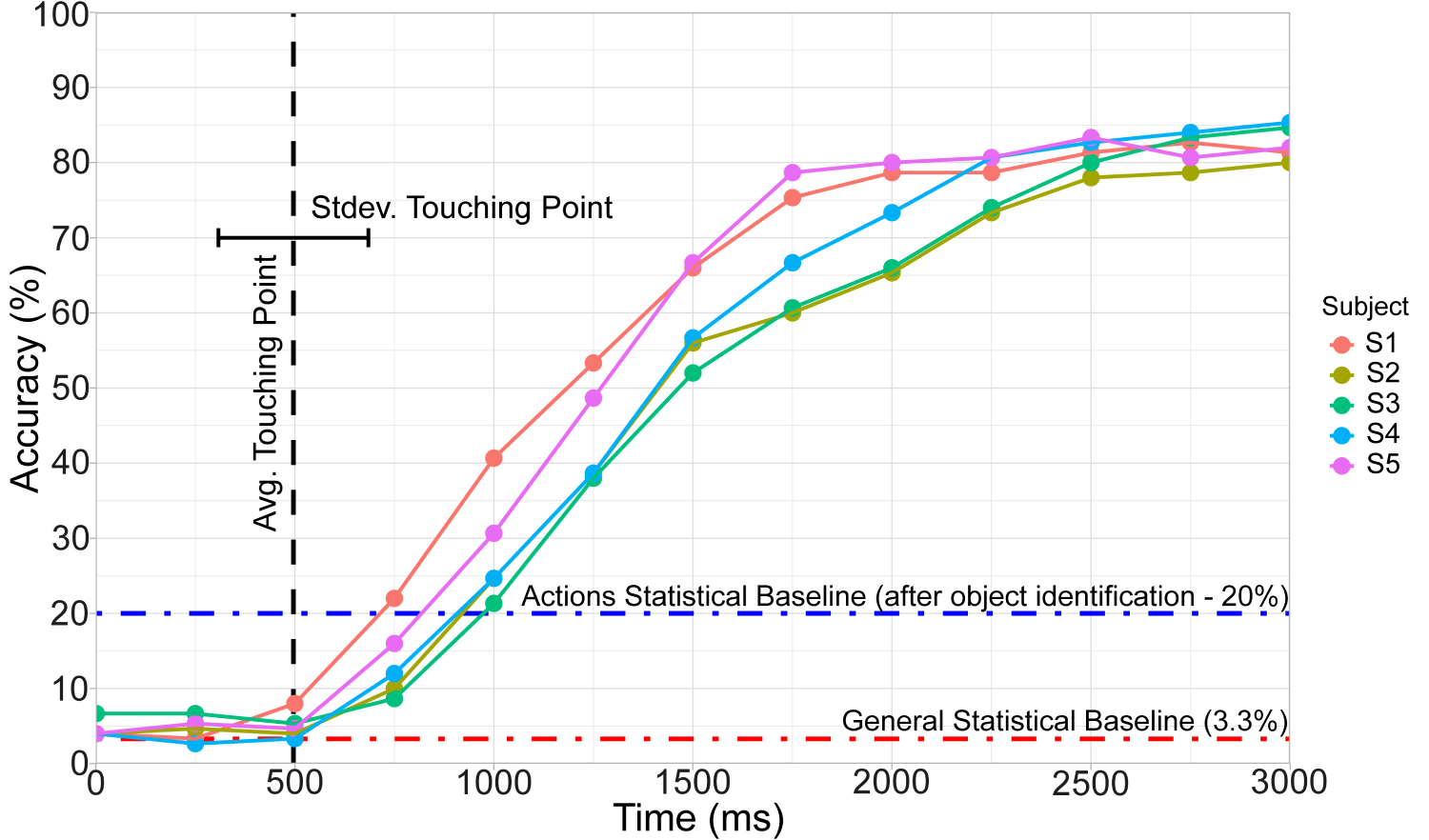}
        \end{minipage}
    \end{center}
    \caption{Left: Accuracy of the \textit{Mobilenet LSTM} for online prediction over time. Right: Accuracy of the \textit{Mobilenet Transformer} for online inference over time. Note how when the object becomes visible just after the hand touches it (touching point), the average accuracy of the models starts significantly peaking. After this point, the manipulation action classification mainly relies on the hand motion and its trajectory.}
    \label{figure:online_prediction_eval}
\end{figure*}



Figure \ref{figure:online_prediction_eval} shows the accuracy of the \textit{Mobilenet LSTM} and the \textit{Mobilenet Transformer} for online event-based prediction over time. 
The x-axis represents the time since the beginning of the sequence. 
Since the event sensor only records motion, the object is not ``seen'' before the hand touches it. At the start of the recordings, the object rests on top of the table and thus does not trigger events. After 250 ms, when the hand starts moving, the predictive models can classify some manipulation actions (accuracy above chance, about 3.3\%).

Classification becomes clearer after the hand touches the object.
Within the [750 ms - 1s] time interval (touching point occurs on average around 500 ms), the models show a confidence of about 40\%.  At this time the subject grasps the object and starts interacting with it. 
After only 1.6 to 1.7~s, the predictive models provide an average accuracy of over 70\%. At this time
the temporal information captures the dynamics of the activity. After 3~s, the overall average accuracy reaches 82\%. 

The two predictive architectures have similar performance in the classification. However, the \textit{Mobilenet Transformer} architecture has better generalization, as observed by its lower inter-subject variability.

 Table \ref{table:eval_predictive_manipulation_actions} summarizes data on the computational complexity. With less than 4 million weight parameters, each model computes around 8.5 Giga floating point operations (GFlops) when processing a two-second event sequence. Note how the \textit{Mobilenet Transformer} network is able to process more than 4700 Time Surfaces per second when running on an RTX 2080 GPU. This means that our efficient event-processing solution has the potential of processing around 150 times more times surfaces during the same time interval. 

\begin{table}[t]
    \centering
    \caption{Time Performance of predictive models}
    \label{table:eval_predictive_manipulation_actions}
    \begin{tabular}{ l r | r r }
        \hline
        \multicolumn{1}{c}{\multirow{2}{*}{DL architecture}} & \multicolumn{1}{c|}{\# Params} & \multicolumn{1}{c}{\multirow{2}{*}{GFlops}} &  \multicolumn{1}{c}{\multirow{2}{*}{TS/s}} \\
        \multicolumn{1}{c}{} & \multicolumn{1}{c|}{(Millions)} \\ 
        \hline
        \Tstrut
        
       Mobilenet LSTM & 3.11 & 8.43 & 4662\\
       \textbf{Mob. Transf. (Ours)} & 3.90 & 8.47 & 4750\\
        \hline
    \end{tabular}
\end{table}

\begin{figure}[t]
    \centering
    \includegraphics[width=0.6\linewidth]{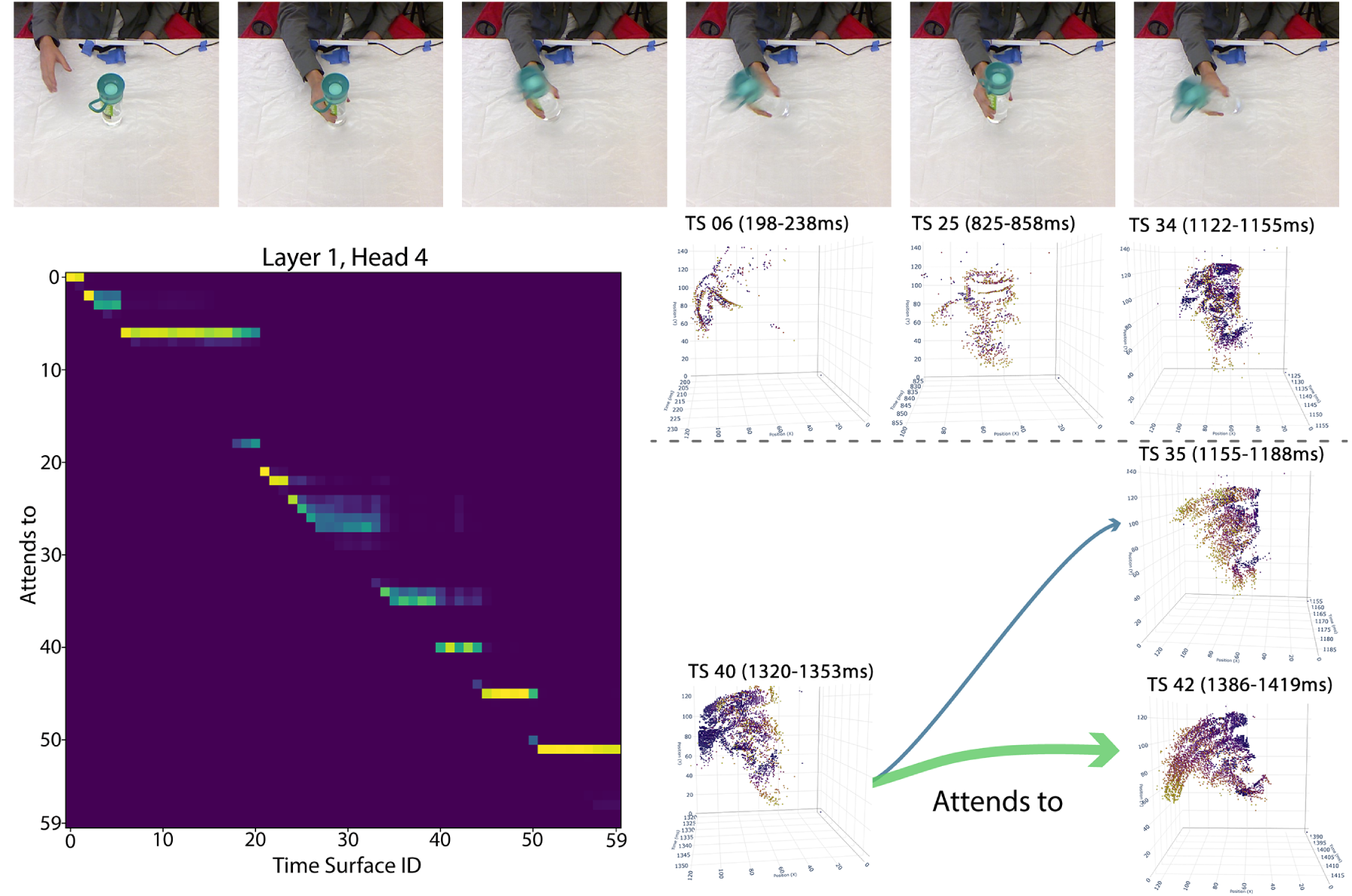}
    \caption{Attention features extracted from a test sample  ``shaking'' a cup. Top: Representation of the action showing several RGB frames from  MAD \cite{fermuller2018predictionurl}. Bottom-left: Attention scores between the events extracted by the attention head 4 of the Transformer block. Bottom-right: On top, time surfaces that were attended the most by subsequent time surfaces. Attention is focused on \textit{TS 40} representing the cup tilted to the left while the shaking action is taking place, the cup coming back to the center position (\textit{TS 35}), and the cup tilted to the right (\textit{TS 42}).}
  \label{figure:attention_cup_shake}
\end{figure}

\begin{figure*}[t]
    \begin{center}
        \begin{minipage}[t]{0.5\linewidth}
	        \centering
 	        \includegraphics[width=\linewidth]{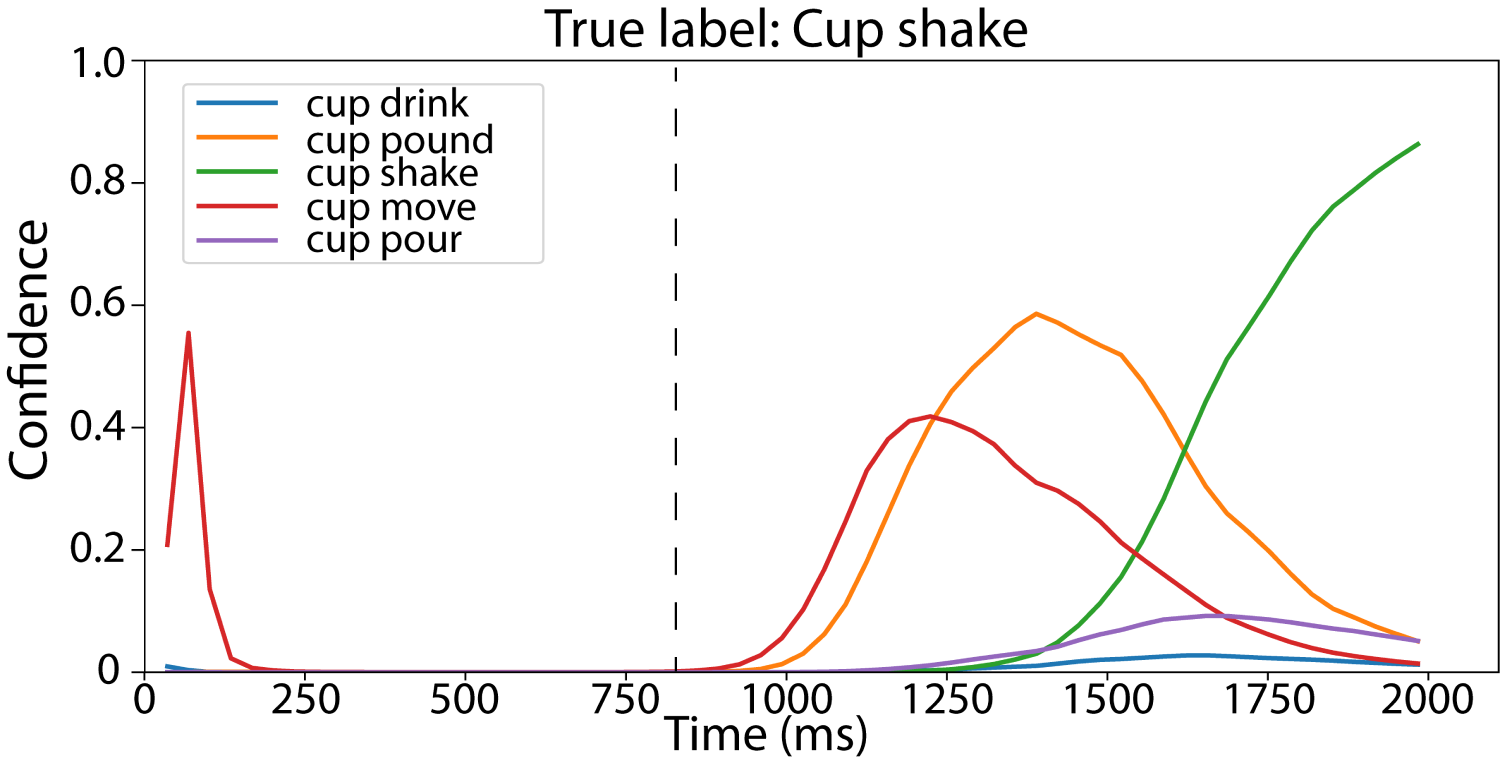}
        \end{minipage}%
        \begin{minipage}[t]{0.5\linewidth}
	        \centering
 	        \includegraphics[width=\linewidth]{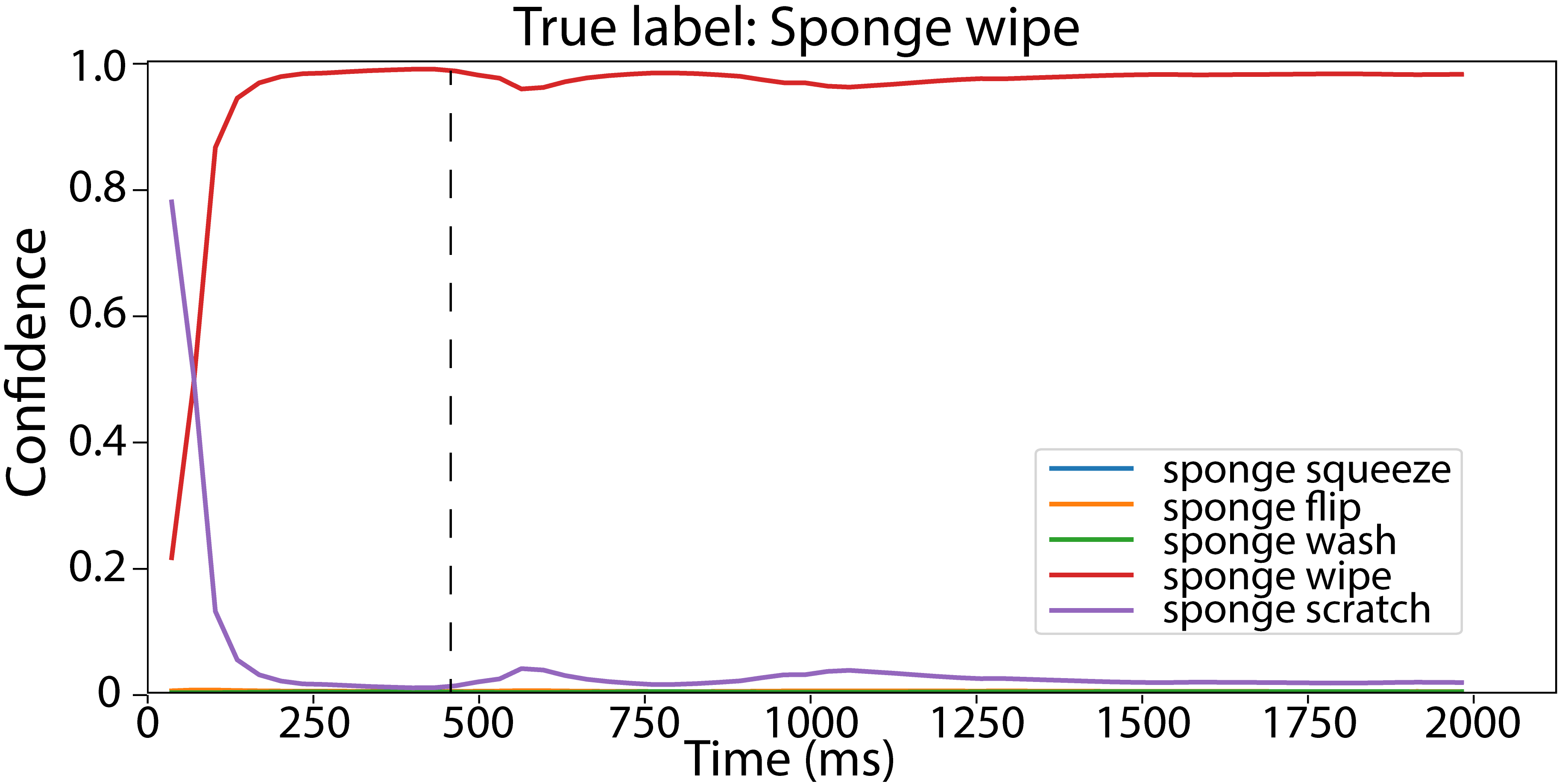}
        \end{minipage}
    \end{center}
    \caption{Left: Output confidence of the \textit{Mobilenet Transformer} for  ``shaking'' a cup. Predictions correspond to the sample shown in Figure \ref{figure:attention_cup_shake}. Right: Output confidence of \textit{Mobilenet Transformer} network processing a sample of ``wiping'' with a sponge. Note how the initial motion of the hand is very enlightening for  ``wiping'', in less than 250~ms the model identifies the correct action with a confidence above 90\%.}
  \label{figure:online_predictions}
\end{figure*}


\subsection{Contribution of causal self-attention}
\label{sec:self_attention}
The Transformer architecture has blocks for causal self-attention. We can analyze these blocks to gain insight into the relationship between attention maps and the prediction of the model.
Attention maps show relationships within the sequence of event temporal surfaces and denote their relevance for classifying a predicted action. 
We 
next analyze the attention scores assigned by the \textit{Transformer block} for a few sample actions. 

\begin{figure*}[t]
    \begin{center}
        \begin{minipage}[t]{0.5\linewidth}
	        \centering
 	        \includegraphics[width=\linewidth]{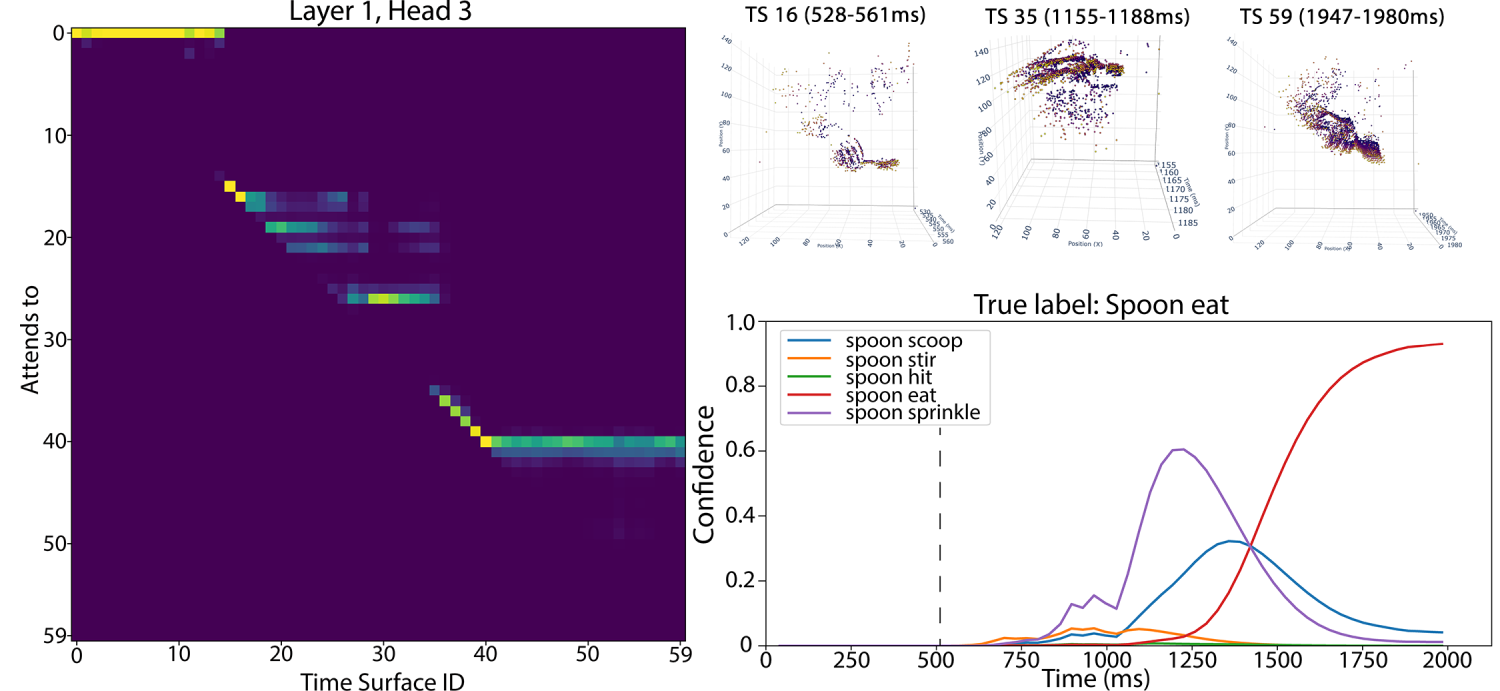}
        \end{minipage}%
        \begin{minipage}[t]{0.5\linewidth}
	        \centering
 	        \includegraphics[width=\linewidth]{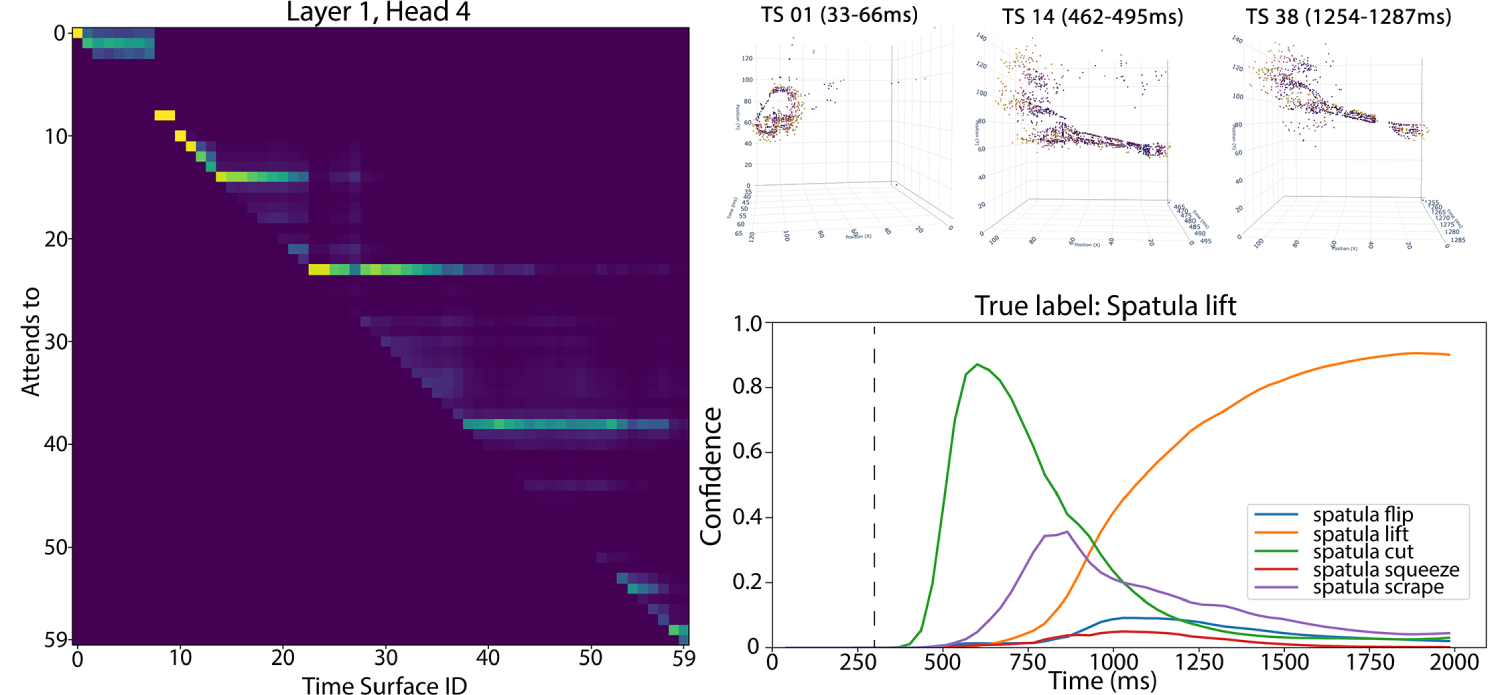}
        \end{minipage}
    \end{center}
    \caption{Left: Attention Maps from attention head 3 and output confidence for a sample of  ``eating'' with a spoon. Right: Attention Maps from attention head 4 and output confidence for online prediction of a sample of ``lifting'' a spatula.}
  \label{figure:spoon_spatula_confidence}
\end{figure*}

\begin{table}[t]
    \centering
    \caption{4-Fold Cross Validation Manipulation Action classification}
    \label{table:eval_manipulation_actions_video}
    \begin{tabular}{l r r | r}
        \hline
        DL architecture & GFlops & Time Performance & Accuracy  \\
        \hline \Tstrut
        Mob. Transf. (Video) & 7.06  & 4750 FPS & $90.02 \pm 4.40$ \\
        Mob. Transf. (Events) & 8.47 & 4750 TS/s & $96.94 \pm 2.50$ \\
        \hline
    \end{tabular}%
\end{table}

\begin{figure}[t]
    \centering
 	\includegraphics[width=0.60\linewidth]{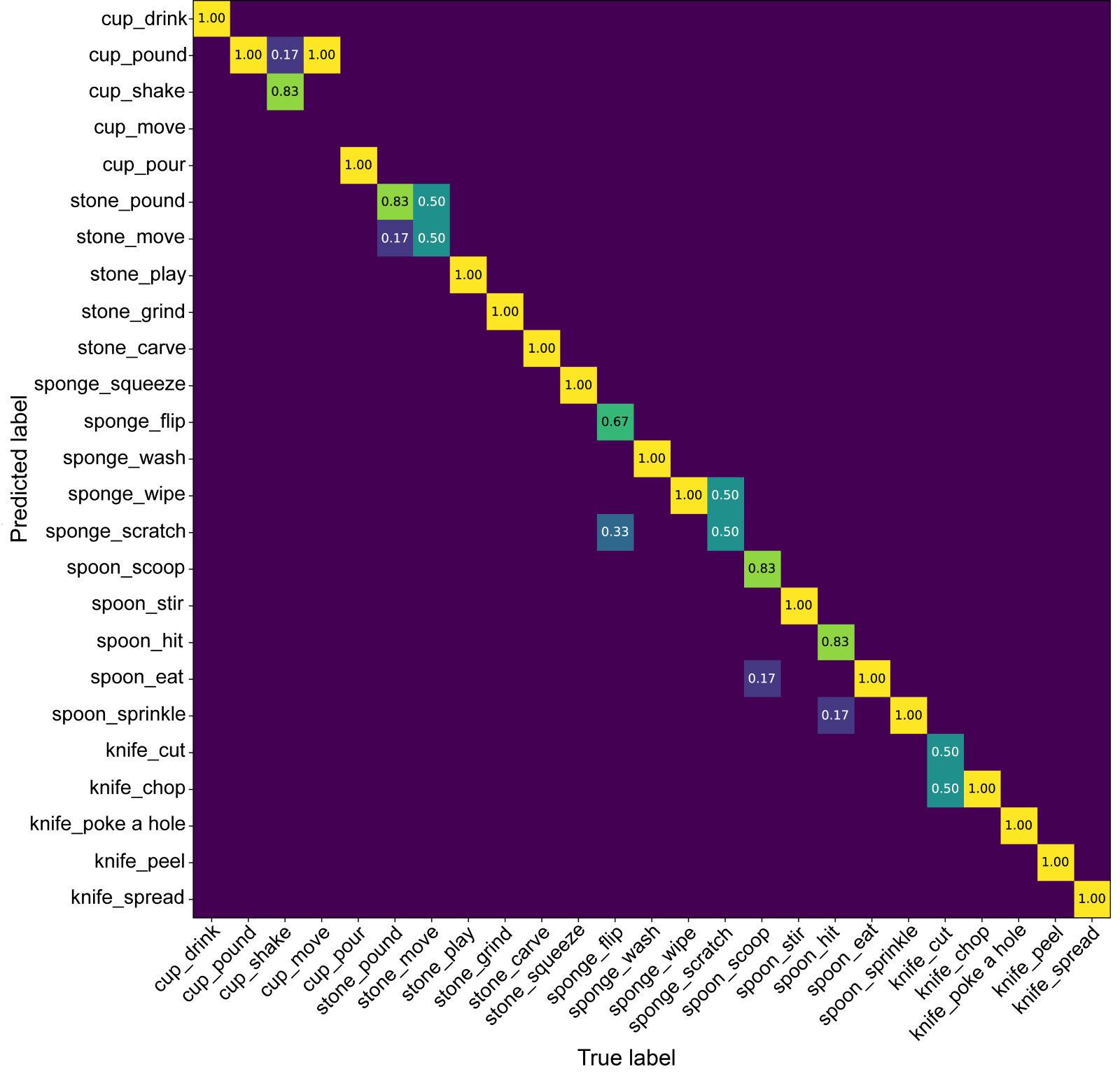}
    \caption{Confusion Matrix for the video-based \textit{Mobilenet Transformer}. There are not classification errors for classes of actions that use different objects but, it presents some errors for actions manipulating a particular object. In this video-based approach, the model tends to focus more on the appearance features than on the motion. For example, the model outputs some misclassifications between actions with the same object, as for the case of the objects cup and stone, with misleading classifications mixing up \textit{pound} and \textit{move}. Contrarily, the event-based solution takes advantage of the intrinsic nature of events that encodes the scene dynamics to obtain more accurate predictions.}
  \label{figure:confusion_matrix_video}
\end{figure}

\begin{figure}[t]
    \centering
 	\includegraphics[width=0.75\linewidth]{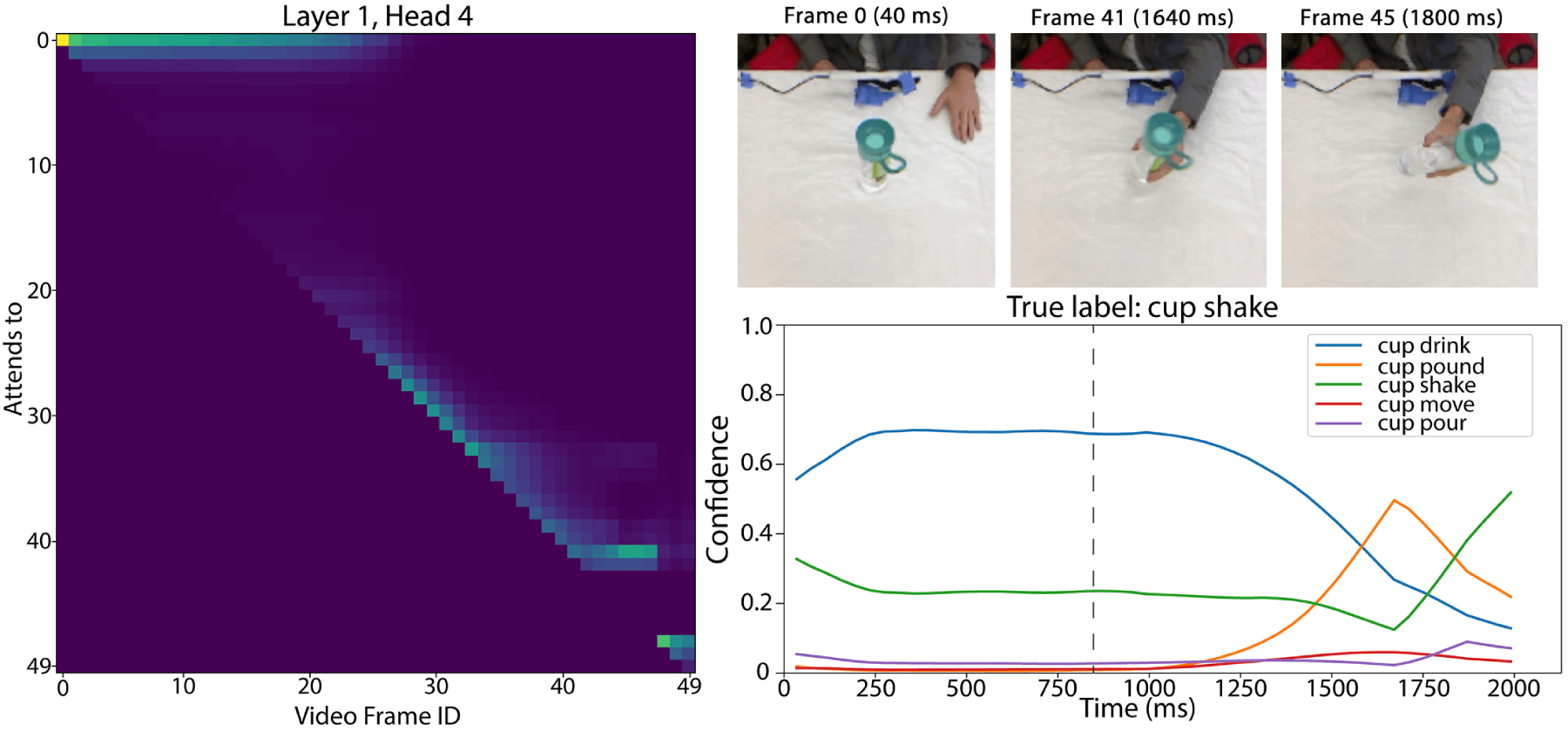}
    \caption{Attention maps from attention head 4 of confidence of video-based \textit{Mobilenet Transformer} for  ``shaking a cup'' (the same video sample that was shown in Figure \ref{figure:attention_cup_shake} for the event-based transformer). Different from the event-based solution, attention is mainly focused on the first frame where only the object is shown. 
    After 2~s the model outputs the correct label ``shake'', although the confidence reaches only around 0.5.
    }
  \label{figure:shaking_cup_video}
\end{figure}

Figure \ref{figure:attention_cup_shake} shows the attention scores assigned by one of the attention heads of the \textit{Mobilenet Transformer} network (Layer 1, Head 4) when analyzing events captured from a subject \textit{shaking a cup} (a \textbf{cyclic} or repetitive action). The left side of the figure shows the attention scores assigned by the model to the different time surfaces. For example, one observes that time surfaces for the initial position of the hand (\textit{TS 06}) and for the hand grasping the cup (\textit{TS 24}) have a higher attention score than others. 
Following  attention is on surfaces 
corresponding to the cyclic motion of shaking the cup left and right.
These surfaces are sufficient for a correct prediction with the neural model. 

Figure \ref{figure:online_predictions}-left shows the online prediction results of the sample in Figure \ref{figure:attention_cup_shake}. Prediction starts after 
the cup is picked at 850~ms, as 
only then the cup becomes visible to the event sensor. 
Then after 1400~ms, the model captures the fast movements of the cup, which are  specific to the action \textit{shaking a cup}, and the 
confidence for \textit{shaking} builds up rapidly  after 2~s. 


 Figure \ref{figure:spoon_spatula_confidence} shows the attention maps and the output confidence for \textit{eating with a spoon} and \textit{lifting a spatula}, two discrete actions. In \textit{eating}  using a spoon (Figure \ref{figure:spoon_spatula_confidence}-left)  the initial position of the hand receives attention (\textit{TS 00}) until the subject grasps the spoon (\textit{TS 16}). Other attended surfaces correspond to the spoon starting going up and reaching the mouth of the subject. In the output confidence chart, after the touching point (around 500~ms) the model identifies the object \textit{spoon} and the confidence for the action \textit{eating} increases when the subject raises the spoon to his/her mouth. When the subject puts down the spoon the confidence increases to 90\%. 
 For  \textit{lifting the spatula} (Figure \ref{figure:spoon_spatula_confidence}-right)  the identifying patterns are due to the hand pose before grasping the object (\textit{TS 01}), the touching point (\textit{TS 14}), and the motion of moving the spatula up. After only 1~s, the predictive \textit{Mobilenet Transformer} is capable of identifying the right action with very high confidence. 

Figure \ref{figure:online_predictions}-right shows the action  \textit{wiping} with a sponge. Here the hand pose before reaching the sponge for 250~ms suffices for the identification of the activity with very high confidence. This is much earlier than the touching point (dashed line around 500~ms).


In summary, the attention mechanism allows us to analyze the patterns most relevant for distinguishing between manipulation actions. Note that causal attention forces the network to pay attention only to past events. 
Some of the clear patterns of attention are:
There is attention to the hand pose at the beginning of the action. 
The touching point is crucial for most 
actions.
Other key features are determined by the action category: for repetitive (\textit{cyclic}) manipulation actions, 
patterns characterizing the cycles get more attention, distributing attention equally among consecutive time surfaces to capture the dynamics of the action during the cycles; for discrete actions, attention is on characteristic movements such as specific gestures or the hand position and pose just before starting the manipulation.

\subsection{Manipulation Action Detection from RGB videos}
We next compare the performance of our event-based Transformer with the corresponding video-based Transformer. Video data provides more information about the static content. However, events provide significantly more accurate information about the motion in the scene.

We train a video-based Transformer on the video-based \textit{MAD} dataset \cite{fermuller2018predictionurl}. The video dataset was collected simultaneously with the event dataset. However, the viewing angles are slightly different, and it only has actions on five of the objects (leaving out the \textit{spatula}).

The computational cost is similar. Referring to  Table~\ref{table:eval_manipulation_actions_video}, we see that the video Transformer has worse performance. The event-based alternative is on average 7.6\% more accurate. The confusion matrix in Figure \ref{figure:confusion_matrix_video} shows how the video model is very accurate at identifying the object that the subject is interacting with. However, it predicts wrong labels when distinguishing between the different actions on the same object.

Additionally, Figure \ref{figure:shaking_cup_video} illustrates attention maps from head 4 and output confidence for the video-based Transformer model processing a sample of ``shaking a cup'' (the same sample that was illustrated in Figure \ref{figure:attention_cup_shake} for the event-based transformer). Note that now, attention is mainly drawn to the first frame which only shows the object on top of the table. Later on, attention is directed to the last frames where the hand is shaking the cup (frames 41 - 45). The model is able to identify with high confidence the object (above 90\% confidence) even before the touching point. This is because, contrarily to the event-based approach, the video-based model captures the object appearance features already from the first frame. However, it takes for the video-based approach almost 2~s to extract features of the action itself to infer the final correct label (\textit{shake}), but with low confidence (and lower than the event-based transformer model). Again, the results seem to show that the event-based model better captures the scene dynamics, outperforming the video-based model.

Table \ref{table:eval_manipulation_actions_video} shows that the two approaches have similar computational complexity. On an RTX 2080 GPU, the event-based model processes up to 4750 time surfaces per second, performing faster than real-time. The event-based approach has the additional advantage of lower latency between predictions at online inference - the time to wait to build a new time surface is 33~ms, but this could easily be reduced. 

~

\section{Conclusions}
This work demonstrates manipulation action prediction using neuromorphic sensors, achieving state-of-the-art results using a Transformer-based neural model. Predictive architectures show better performance in the ablation study, additionally enabling online inference. 
Online inference helps to leverage the asynchronous nature of event data and to reduce the latency of inferring new predictions. Moreover, online continuous prediction reduces latencies by up to 2-3~s in our experiments compared to classification models. 
Significant latency reduction is crucial for many applications, such as robotic collaboration or gesture recognition. 


Another key aspect of our attentional model is the interpretability of the prediction decision process. In our case, a by-product of the model is the taxonomy of actions that naturally emerges from the learning process: \textit{cyclic} actions that show a repetitive pattern such as \textit{shaking a cup} or \textit{poking a hole}, and \textit{discrete} actions that are done only once such as \textit{eating with a spoon}. The analysis shows that attention for \textit{cyclic} actions is drawn by the time surfaces that correspond to the starting and end points of the cycles. For \textit{discrete} actions, the focus of attention is on the hand pose before the manipulation and the trajectory of the hand + object while performing the action. A comparison showed that video-based approaches extract features that are mostly appearance-based while event-driven approaches extract dynamic features of the actions. This makes event-driven solutions better candidates for classifying actions with subtle dynamic differences, for example, different manipulation actions on the same objects. 

We conclude that the Transformer, because of its ability to perform online prediction and because of its good performance, is well suited for identifying manipulation actions from event-driven sensors. 
Compared to the video-based model, the event-driven model has  better performance with shorter latencies, 
and infers in real-time.      


\section*{Acknowledgements}

\noindent This work was supported by the Spanish National Grant PID2019-109434RA-I00/ SRA (State Research Agency /10.13039/501100011033). We acknowledge the Telluride Neuromorphic Cognition Engineering Workshop (\url{http://www.ine-web.org}), supported by NSF grant  OISE 2020624 for the fruitful discussions on neuromorphic cognition and their participants for helping with the recording of the dataset.


\bibliographystyle{unsrt}
\bibliography{bibliography}







\end{document}